%% file: paper.tex
\documentclass{article} 
\usepackage{iclr2017_conference,times}
\usepackage{hyperref}
\usepackage{url}
\usepackage{enumitem}
\usepackage[utf8]{inputenc}
\usepackage[T1]{fontenc}
\input{defs} 

\title{N-gram Language Modeling using Recurrent Neural Network Estimation}

\author{
Ciprian Chelba, Mohammad Norouzi, Samy Bengio\\
Google\\
~\texttt{\{ciprianchelba,mnorouzi,bengio\}@google.com}
}

\iclrfinalcopy

\begin{document}

\maketitle

\input{abs}
\input{intro}
\input{method}
\input{expts}
\input{conc}

\section*{Acknowledgments}

We would like to thank Oriol Vinyals and Rafał Józefowicz for support with the baseline implementation of LSTM LMs for UPenn Treebank in~\cite{ptb_lm_tutorial} and One Billion Words Benchmark in~\cite{rafal:one-billion-wds}, respectively. We would also like to thank Maxim Krikun for thorough code reviews and useful discussions.

\bibliography{paper}
\bibliographystyle{iclr2017_conference}

\end{document}

%% file: defs.tex
\usepackage{amsthm}
\usepackage{amsmath}
\usepackage{amssymb}
\usepackage{color}
\usepackage{xspace}
\usepackage{pgfplots, pgfplotstable}

\newcommand{\comment}[1]{}

\def\int{\mathrm{int}}

%% file: abs.tex
\begin{abstract}

  We investigate the effective memory depth of RNN models by using them for $n$-gram language model (LM) smoothing.

  Experiments on a small corpus (UPenn Treebank, one million words of training data and 10k vocabulary) have found the LSTM cell with dropout to be the best model for encoding the $n$-gram state when compared with feed-forward and vanilla RNN models. When preserving the sentence independence assumption the LSTM $n$-gram matches the LSTM LM performance for $n=9$ and slightly outperforms it for $n=13$. When allowing dependencies across sentence boundaries, the LSTM $13$-gram almost matches the perplexity of the unlimited history LSTM LM.

  LSTM $n$-gram smoothing also has the desirable property of improving with increasing $n$-gram order, unlike the Katz or Kneser-Ney back-off estimators. Using multinomial distributions as targets in training instead of the usual one-hot target is only slightly beneficial for low $n$-gram orders.

  Experiments on the One Billion Words benchmark show that the results hold at larger scale: while LSTM smoothing for short $n$-gram contexts does not provide significant advantages over classic N-gram models, it becomes effective with long contexts ($n > 5$); depending on the task and amount of data it can match fully recurrent LSTM models at about $n=13$. This may have implications when modeling short-format text, e.g. voice search/query LMs.

  Building LSTM $n$-gram LMs may be appealing for some practical situations: the state in a $n$-gram LM can be succinctly represented with $(n-1)*4$ bytes storing the identity of the words in the context and batches of $n$-gram contexts can be processed in parallel. On the downside, the $n$-gram context encoding computed by the LSTM is discarded, making the model more expensive than a regular recurrent LSTM LM.

\end{abstract}

%% file: intro.tex
\section{Introduction}

A statistical language model (LM) estimates the prior probability values $P(W)$ for
strings of words $W$ in a vocabulary ${\cal V}$ whose size is usually in the
tens or hundreds of thousands. Typically the string $W$ is broken
into sentences, or other segments such as utterances in automatic speech
recognition which are assumed to be conditionally independent; the independence
assumption has certain advantages in practice but is not strictly necessary.

Applying the chain rule to a sentence $W=w_1,w_2,\ldots,w_n$ we get:
\begin{eqnarray}
  \label{intro:bayes}
  P(W)=\prod_{k=1}^nP(w_k|w_1,w_2,\ldots,w_{k-1}) 
\end{eqnarray}

Since the parameter space of $P(w_k|w_1,w_2,\ldots,w_{k-1})$ is too large, the language model is forced to put the
\emph{context} $W_{k-1}=w_1,w_2,\ldots,w_{k-1}$ into an \emph{equivalence class}
determined by a function $\Phi(W_{k-1})$. As a result:
\begin{equation}
  \label{e1}P(W)\cong\prod_{k=1}^nP(w_k|\Phi (W_{k-1})) 
\end{equation}

Research in language modeling consists of finding appropriate
equivalence classifiers $\Phi$ and methods to estimate
$P(w_k|\Phi(W_{k-1}))$.

\subsection{Perplexity as a Measure of Language Model Quality}

A commonly used quality measure for a given model $M$ is related to
the entropy of the underlying source and was introduced under
the name of \emph{perplexity} (PPL)~\cite{jelinek97}:
\begin{eqnarray}
\label{basic_lm:ppl}
PPL(W,M) = exp(-\frac{1}{N} \sum_{k=1}^{N}\ln{[P_M(w_k|W_{k-1})]}) 
\end{eqnarray}
To give intuitive meaning to perplexity, it represents the average
number of guesses the model needs to make in order to ascertain the identity of
the next word, when running over the test word string $W = w_1 \ldots w_N$
from left to right. It can be easily shown that the perplexity of a language model
that uses the uniform probability distribution over words in the
vocabulary ${\cal V}$ equals the size of the vocabulary; a good
language model should of course have lower perplexity, and thus the
vocabulary size is an upper bound on the perplexity of a given
language model.

Very likely, not all words in the test data are
part of the language model vocabulary. It is common practice to map
all words that are out-of-vocabulary to a distinguished \emph{unknown word}
symbol, and report the out-of-vocabulary (OOV) rate on test data---the
rate at which one encounters OOV words in the test sequence $W$---as
yet another language model performance metric besides
perplexity. Usually the unknown word is assumed to be part of the
language model vocabulary---\emph{open vocabulary} language models---and
its occurrences are counted in the language model perplexity
calculation in Eq.~(\ref{basic_lm:ppl}). A situation less common in
practice is that of \emph{closed vocabulary} language models where all
words in the test data will always be part of the vocabulary ${\cal  V}$.

\subsection{Smoothing}

Since the language model is meant to assign non-zero probability to unseen
strings of words (or equivalently, ensure that the cross-entropy of
the model over an arbitrary test string is not infinite), a desirable
property is that: 
\begin{eqnarray}
P(w_k|\Phi (W_{k-1})) > \epsilon > 0, \forall w_k, W_{k-1}  \label{eq:smoothing}
\end{eqnarray}
also known as the \emph{smoothing} requirement.

There are currently two dominant approaches for building LMs:

\subsubsection{$n$-gram Language Models}

The most widespread paradigm in language modeling makes a Markov assumption and uses
the \emph{$(n-1)$-gram} equivalence classification, that is, defines%
\begin{equation}
  \label{e2}\Phi (W_{k-1})\doteq w_{k-n+1},w_{k-n+2},\ldots,w_{k-1} = h
\end{equation}

A large body of work has accumulated over the years on various smoothing methods for $n$-gram
LMs. The two most popular smoothing techniques are probably~\cite{Kneser:95a}
and~\cite{katz:back_off}, both making use of \emph{back-off} to balance the specificity
of long contexts with the reliability of estimates in shorter $n$-gram contexts.
\cite{Goodman:01b} provides an excellent overview that is highly recommended to any practitioner
of language modeling.

Approaches that depart from the nested features used in back-off $n$-gram LMs have shown
excellent results at the cost of increasing the number of features and
parameters stored by the model, e.g.~\cite{TACL561}.

\subsubsection{Neural Language Models}

Neural networks (NNLM) have emerged in recent years as an alternative to estimating and
storing $n$-gram LMs. Words (or some other modeling unit) are
represented using an embedding vector $E(w) \in \mathbb{R}^d$.
A simple NNLM architecture makes the Markov assumption and feeds the concatenated
embedding vectors for the words in the $n$-gram context to one or more layers each
consisting of an affine transform followed by a non-linearity (typically $\tanh$);
the output of the last such layer is then fed to the output layer consisting again of an affine
transform but this time followed by an exponential non-linearity that is
normalized to guarantee a proper probability over the vocabulary. This is commonly
named a feed-forward architecture for an $n$-gram LM (FF-NNLM), first introduced by~\cite{nnlm:2001:nips}.

An alternative is the recurrent NNLM architecture that feeds the embedding of each word
$E(w_k)$ one at a time, advancing the state $S \in \mathbb{R}^s$ of a \emph{recurrent cell}
and producing a new output $U \in \mathbb{R}^u$:
\begin{eqnarray}
  [S_{k}, U_{k}] & = & RNN(S_{k-1}, E(w_{k}))\label{e3}\\
  S_{0} & = & 0 \nonumber
\end{eqnarray}

This provides a representation for the context $W_{k-1}$ that can be directly plugged into Eq.~\ref{e1}:
\begin{eqnarray}
  \Phi(W_{k-1}) = U_{k-1}(W_{k-1})
\end{eqnarray}

Similar to the FF-NNLM architecture, the output $U$ of the recurrent cell is then fed to a soft-max layer consisting of an affine transform $O$ followed by an exponential non-linearity properly normalized over the vocabulary.

The recurrent cell $RNN(\cdot)$ can consist of one or more simple affine/non-linearity layers, often called a \emph{vanilla RNN} architecture, see~\cite{mikolov2010recurrent}. The \emph{LSTM} cell due to~\cite{lstm} has proven very effective at modeling long range dependencies and has become the state-of-the-art architecture for language modeling using RNNs, see~\cite{DBLP:journals/corr/JozefowiczVSSW16}.

In this work we approximate unlimited history (R)NN models with $n$-gram models in an attempt to identify the order $n$ at which they become equivalent from a perplexity point of view. This is a promising direction in a few ways:
\begin{itemize}
\item the training data can be reduced to $n$-gram sufficient statistics, and the target distribution presented to the NN $n$-gram LM in a given context can be a multinomial pmf instead of the one-hot encoding used in on-line training for (R)NN LMs;
\item unlike many LSTM LM implementations, back-propagation through time for the LSTM $n$-gram need not be truncated at the begining of segments used to batch the training data;
\item the state in a $n$-gram LM can be succinctly represented with $(n-1)*4$ bytes storing the identity of the words in the context; this is in stark contrast with the state $S \in \mathbb{R}^s$ for an RNN LM, where $s=1024$ or higher, making the $n$-gram LM much easier to use in decoders such as for ASR/SMT;
\item similar to~\cite{brants-EtAl:2007:EMNLP-CoNLL2007}, batches of $n$-gram contexts can be processed in parallel to estimate a \emph{sharded} (R)NN $n$-gram model; this is particularly attractive because it allows scaling both the amount of training data and the NNLM size significantly (100X).
\end{itemize}

%% file: method.tex
\section{Method}
\label{method}

As mentioned in the previous section, the Markov assumption made by $n$-gram models allows us to present to the NN \emph{multinomial} training targets specifying the full distribution in a given $n$-gram context instead of the usual \emph{one-hot} target specifying the predicted word occurring in a given context instance. In addition, when using \emph{multinomial} targets we can either weight each training sample by the context count or simply present each context token encountered in the training data along with the conditional multinomial pmf computed from the entire training set.

We thus have three main training regimes:
\begin{itemize}
\item context-weighted multinomial targets
\item multinomial targets (context count $count(h) = 1$)
\item one-hot targets (context count $count(h) = 1$, word count $count(h,w) = 1$)
\end{itemize}

The loss function optimized in training is the cross-entropy between the model pmf $P(w|h; \theta)$ in some $n$-gram context $h$ and the relative frequency $f(w|h; \mathcal{T})$ in the training data $\mathcal{T}$ (or development $\mathcal{D}$, or test data $\mathcal{E}$) is computed as:
\begin{equation}
  \label{x-entropy}H(P, \mathcal{T}) = - 1/T \sum_h count(h) \sum_w f(w|h; \mathcal{T}) \log P(w|h; \theta)
\end{equation}
where $T$ is the length of the training data $\mathcal{T}$ and $P(\cdot; \theta)$ is the $n$-gram model being evaluated/trained as parameterized by $\theta$.

The baseline back-off $n$-gram models (Katz, interpolated Kneser-Ney) are trained by making a sentence independence assumption. As a result $n$-gram contexts at the beginning of the sentence are padded to the left to reach the full context length. The same $n$-gram counting strategy is used when preparing the data for the various NN $n$-gram LMs that we experimented with. Since RNN LMs are normally trained and evaluated without making this independence assumption by passing the LM state across sentence boundaries, we also evaluated the impact of resetting the RNN LM state at sentence beginning.

We next detail the various flavors or NN LM implementations we experimented with.

For all NN LMs we represent context words using an embedding vector $E(w) \in \mathbb{R}^d$. Unless otherwise stated, all models are trained to minimize the cross-entropy on training data in Eq.~(\ref{x-entropy}), using Adagrad~(\cite{Duchi:2011:ASM:1953048.2021068}) and gradient norm clipping~(\cite{DBLP:journals/corr/abs-1211-5063}); the model parameters are initialized by sampling from a truncated normal distribution of zero mean and a given standard deviation.

Training proceeds for a fixed number of epochs for every given point on the grid of hyper-parameters explored for a given model type; the best performing model (parameter values) on development data $\mathcal{D}$ is retained as the final one to be evaluated on test data $\mathcal{E}$ in order to report the model perplexity.

All models were implemented using TensorFlow, see~\cite{tensorflow2015-whitepaper}.

\subsection{Feed Forward $n$-gram LM}

Each word $w$ in the $n$-gram context $h=w_{k-n+1} \ldots w_{k-1}$ is embedded using the mapping $E(w)$; the resulting vectors are concatenated to form a $d \cdot (n-1)$ dimensional vector that is first fed into a dropout layer~\cite{JMLR:v15:srivastava14a} and then into an affine layer followed by a $\tanh$ non-linearity. The output of this so-called ``hidden'' layer is again fed into a dropout layer and then followed by an affine layer $O$ whose output is of the same dimensionality as the vocabulary. An exponential ``soft-max'' layer converts the activations produced by the last affine layer into probabilities over the vocabulary.

To summarize:
\begin{eqnarray}
  X & = & concat(E(w_{k-n+1}), \ldots,  E(w_{k-1}))\nonumber\\
  D(X) & = & dropout(X; P_{keep})\nonumber\\
  Y & = & \tanh(H \cdot D(X) + H_{bias})\nonumber\\
  D(Y) & = & dropout(Y; P_{keep})\nonumber\\
  P(\cdot|w_{k-n+1} \ldots w_{k-1}) & = & \exp(O \cdot D(Y) + O_{bias})\label{feed-fwd}
\end{eqnarray}

The parameters of the model are the embedding matrix $E \in \mathbb{R}^{d \times V}$, the keep probability for dropout layers $P_{keep}$, the affine input layer parameterized by $H \in \mathbb{R}^{s \times (n-1) \cdot d}$, $H_{bias} \in \mathbb{R}^{s}$ and the output one parameterized by $O \in \mathbb{R}^{V \times s}$, $O_{bias} \in \mathbb{R}^{V}$.

The hyper-parameters controlling the training are: number of training epochs, $n$-gram order, dimensionality of the model parameters $d, s$, keep probability value, gradient norm clipping value, standard deviation for the initializer and the Adagrad learning rate and initial accumulator value.

\subsection{``Vanilla'' Recurrent $n$-gram LM}

Each word $w$ in the $n$-gram context $h=w_{k-n+1} \ldots w_{k-1}$ is embedded using the mapping $E(w)$ followed by dropout and then fed in left-to-right order into the RNN cell in Eq.~(\ref{e3}). The final output of the RNN cell is then fed first into a dropout layer and then into an affine layer followed by exponential ``soft-max''.

Assuming that we encode the context $h=w_{k-n+1} \ldots w_{k-1}$ with an RNN cell defined as follows (using the running index $l = k-n+1 \ldots k$ to traverse the context and a dropout layer on the embedding $D(E(w_{l})) = dropout(E(w_{l}), P_{keep})$):
\begin{eqnarray}
  [S_{l}, U_{l}] & = & \tanh(R \cdot [S_{l-1}, D(E(w_{l}))] + R_{bias})\nonumber\\
  S_{k-n} & = & 0 \label{rnn-cell}
\end{eqnarray}

we pick the last output $U_{k-1}$ and feed it into a dropout layer followed by an affine layer and soft-max output:
\begin{eqnarray}
  D(U_{k-1}) & = & dropout(U_{k-1}; P_{keep})\nonumber\\
  P(\cdot|w_{k-n+1} \ldots w_{k-1}) & = & \exp(O \cdot D(U_{k-1}) + O_{bias})\label{rnn-lm}
\end{eqnarray}

The parameters of the model are the embedding matrix $E \in \mathbb{R}^{d \times V}$, the keep probability for dropout layers $P_{keep}$, the RNN affine layer parameterized by $R \in \mathbb{R}^{(d + s) \times 2 \cdot s)}$ and $R_{bias} \in \mathbb{R}^{2 \cdot s}$ and the output one parameterized by $O \in \mathbb{R}^{V \times s}$ and $O_{bias} \in \mathbb{R}^{V}$. Note that we choose to use the same dimensionality $s$ for both $S,U \in \mathbb{R}^{s}$.

The hyper-parameters controlling the training are the same as in the previous section.

\subsection{LSTM Recurrent $n$-gram LM}

Finally, we replace the ``vanilla'' RNN cell defined above with a multilayer LSTM cell with dropout.
Since this was the most effective model, we experimented with a few options:
\begin{itemize}
\item forward context encoding: context words $h=w_{k-n+1} \ldots w_{k-1}$ are fed in left-to-right order in the LSTM cell; the LSTM cell output after the last context word $w_{k-1}$ is then fed into the output layer;
\item reverse context encoding: context words $h=w_{k-n+1} \ldots w_{k-1}$ are fed in left-to-right order in the LSTM cell; the LSTM cell output after the first context word $w_{k-n+1}$ is then fed into the output layer;
\item stacked output for either of the above: we concatenate the output vectors along the way and feed that into the output layer;
\item bidirectional context encoding: we encode the context twice, forward and reverse order respectively, using two separate LSTM cells; the two outputs are then concatenated and fed to the output layer;
\item forward context encoding with incremental loss with/out exponential decay as a function of the context length
\end{itemize}

The last item above deserves a more detailed explanation. It is possible that the LSTM encoder would benefit from incremental error back-propagation along the $n$-gram context instead of just one back-propagation step at the end of the context. As such, we modify the loss function to be the cumulative cross-entropy between the relative frequency and the model output distribution \emph{at each step in the for loop feeding the $n$-gram context into the LSTM cell instead of just the last one}. Thus amounts to targetting a mix of $1 \ldots n$-gram target distributions; to have better control over the contribution of different $n$-gram orders to the loss function, we weigh each loss function by an exponential term $exp(-decay \cdot (n-1-l))$. The $decay > 0$ value controls how fast the contribution to the loss function from lower $n$-gram orders decays; note that the highest order $l=n-1$ has weight $1.0$ so a very large value $decay = \infty$ restores the regular training loss function. For this training regime we only implemented one-hot targets: the amount of data that needs to be fed to the TensorFlow graph would increase significantly for incremental multinomial targets. 

The hyper-parameters controlling the training are: number of training epochs, $n$-gram order, embedding dimensionality $d$, LSTM cell output dimensionality $s$ and number of layers, keep probability value, gradient norm clipping value, standard deviation for the initializer. To match the fully recurrent LSTM LM implemented by the UPenn Treebank TensorFlow tutorial, we estimated all of our LSTM $n$-gram models using gradient descent with variable learning rate: initially the learning rate is constant for a few iterations after which it follows a linear decay schedule. The hyper-parameters controlling this schedule were not optimized but rather we used the same values as in the RNN LM tutorial provided with~\cite{ptb_lm_tutorial} or the implementation in~\cite{rafal:one-billion-wds}, respectively.

Perhaps a bit of a technicality but it is worth pointing out a major difference between error back-propagation through time (BPTT) as implemented in either of the above and the error back-propagation in the LSTM/RNN $n$-gram LM:~\cite{ptb_lm_tutorial} and~\cite{rafal:one-billion-wds} implement BPTT by \emph{segmenting} the training data into non-overlapping segments (of length 35 or 20, respectively)\footnote{We have evaluated the impact of reducing the segment length dramatically, e.g. 4 instead of 35. Much to our surprise, the LSTM PPL increased modestly, from 84 to 88, see the before last row in Table~\ref{table:baseline}; for the One Billion Words experiments using a segment of length 5 did not change PPL at all.}. The error BPTT does not cross the left boundary of such segments, whereas the LSTM state is of course copied forward. As a result, the first word in a segment is not really contributing to training, and the immediately following ones have a limited effect. This is in contrast to error back-propagation for the LSTM/RNN $n$-gram LM: the $n$-gram window slides over the training/test data, and error back-propagation covers the entire $n$-gram context; the LSTM cell state and output computed for a given $n$-gram context are discarded once the output distribution is computed.

%% file: expts.tex
\section{Experiments}

\subsection{UPenn Treebank Corpus}
\label{upenn_exps}

For our initial set of experiments we used the same data set as in~\cite{ptb_lm_tutorial}, with exactly the same training/validation/test set partition and vocabulary. The training data consists of about one million words, and the vocabulary contains ten thousand words; the validation/test data contains 73760/82430 words, respectively (including the end-of-sentence token). The out-of-vocabulary rate on validation/test data is 5.0/5.8\%, respectively.

As an initial batch of experiments we trained and evaluated back-off $n$-gram models using Katz and interpolated Kneser-Ney smoothing. We also used the medium setting in~\cite{ptb_lm_tutorial} as an LSTM/RNN LM baseline; since the baseline $n$-gram models are trained under a sentence independence assumption, we also ran the LSTM/RNN LM baseline by resetting the LSTM state at each sentence beginning. The results are presented in Table~\ref{table:baseline}.

\begin{table}[h!]
\centering
\begin{tabular}{|l|r|r|}
\hline
Model & Order & Test PPL \\\hline\hline
\multicolumn{3}{|l|}{\bf{$n$-gram, baseline}}\\\cline{1-1}
Katz, back-off                    & 5 & 167 \\
Katz, back-off                    & 9 & 182 \\
Interpolated Kneser-Ney, back-off & 5 & 143 \\
Interpolated Kneser-Ney, back-off & 9 & 143 \\\hline
\multicolumn{3}{|l|}{\bf{LSTM RNN, baseline}}\\\cline{1-1}
LSTM (medium setting) & reset state at \verb+<S>+  & 95 \\
LSTM (medium setting) & $\infty$  & 84 \\\hline
\end{tabular}
\caption{UPenn Treebank: baseline back-off $n$-gram and LSTM perplexity values.}
\label{table:baseline}
\end{table}

As expected Kneser-Ney (KN) is better than Katz, and it does not improve with the $n$-gram order past a certain value, in this case $n=5$. This behavior is due to the fact that the $n$-gram hit ratio on test data (number of test $n$-grams that were observed in training) decreases dramatically with the $n$-gram order: the percentage of covered $n$-grams\footnote{For the hit ratio calculation the $n$-grams are not padded to the left of sentence beginning; if we are to count hit ratios using padded $n$-grams, the values are: 100, 81, 44.7, 24.0, 16.5, 13.7, 12.5, 11.8, 11.5, respectively.} for $n=1 \ldots 9$ is 100, 81, 42, 18, 8.6, 5.0, 3.3, 2.5, 2.0, respectively.

The medium setting for the LSTM LM in~\cite{ptb_lm_tutorial} performs significantly better than the KN baseline. Resetting the state at sentence beginning degrades PPL significantly by 13\% relative.

\begin{table}[h!]
\centering
\begin{tabular}{|l|r|r|r|}
\hline
Model & Order & \multicolumn{2}{c|}{Test PPL}\\\hline
\multicolumn{2}{|l|}{Training Target} & multinomial & one-hot \\\hline\hline
\multicolumn{4}{|l|}{\bf{$n$-gram, baseline}}\\\cline{1-1}
Interpolated Kneser-Ney, back-off & 5 & \multicolumn{2}{c|}{143}\\\hline
\multicolumn{4}{|l|}{\bf{Feed-fwd $n$-gram}}\\\cline{1-1}
Feed-fwd $n$-gram                    &  5 & 127 & 128 \\
Feed-fwd $n$-gram                    &  9 & 125 & 126 \\
Feed-fwd $n$-gram                    & 13 & 125 & 127 \\\hline
\multicolumn{4}{|l|}{\bf{``Vanilla'' RNN $n$-gram}}\\\cline{1-1}
RNN $n$-gram                         &  9 & 127 & 131 \\\hline
\multicolumn{4}{|l|}{\bf{LSTM RNN $n$-gram}}\\\cline{1-1}
LSTM $n$-gram, forward context encoding &  5 & 103 & 106 \\
LSTM $n$-gram, forward context encoding &  9 &  94 & 93 \\
LSTM $n$-gram, forward context encoding & 13 &  91 & 90 \\\hline
LSTM $n$-gram, reversed context encoding &  9 & 102 & 107 \\
LSTM $n$-gram, bidirectional context encoding &  9 & 100 & 102 \\
incremental LSTM $n$-gram with decay,$decay=2.0$ &  13 & --- & 91 \\\hline
\multicolumn{4}{|l|}{\bf{LSTM RNN, baseline}}\\\cline{1-1}
LSTM (medium setting) & reset at \verb+<S>+  & --- & 95 \\\hline
\end{tabular}
\caption{UPenn Treebank: perplexity values for neural network smoothed $n$-gram LM.}
\label{table:nn-ngram}
\end{table}

We then trained and evaluated various NN-smoothed $n$-gram LMs, as described in Section~\ref{method}. The results are presented in Table~\ref{table:nn-ngram}. The best model among the ones considered is by far the LSTM $n$-gram. The most significant experimental result is that the LSTM $n$-gram can match and even outperform the fully recurrent LSTM LM as we increase the order $n$: $n=9$ matches the LSTM LM performance, decreasing the LM perplexity by 34\% relative over the Kneser-Ney baseline. LSTM $n$-gram smoothing also has the desirable property of improving with the $n$-gram order, unlike the Katz or Kneser-Ney back-off estimators, which can be credited to better feature extraction from the $n$-gram context.

Multinomial targets can slightly outperform the one-hot ones although the difference is shrinking as we increase the $n$-gram order. Weighting the contribution of each context to the loss function by its count did not work; we suspect this is because on-line training does not work well with the Zipf distribution on context counts.

Among the various flavors of LSTM models we experimented with, the forward context encoding performs best. The incremental LSTM $n$-gram with a fairly large decay ($decay=2.0$) is slightly better but we do not consider the difference to be statistically significant (it also entails significantly more computation, we need to perform $n-1$ back-propagation steps for each input $n$-gram).

To compare with the LSTM RNN LM that does not reset state at sentence beginning we also trained LSTM $n$-gram models (forward context encoding only) that straddle the sentence beginning. The results are presented in Table~\ref{table:straddling-nn-ngram}. Again, we notice that for a large enough order the LSTM $n$-gram LM comes very close to matching the fully recurrent LSTM baseline.

\begin{table}[h!]
\centering
\begin{tabular}{|l|r|r|r|}
\hline
Model & Order & \multicolumn{2}{c|}{Test PPL}\\\hline
\multicolumn{2}{|l|}{Training Target} & multinomial & one-hot \\\hline\hline
\multicolumn{4}{|l|}{\bf{LSTM RNN $n$-gram}}\\\cline{1-1}
LSTM $n$-gram, forward context encoding, straddling \verb+<S>+ &  5 & 102 & 104 \\
LSTM $n$-gram, forward context encoding, straddling \verb+<S>+ &  9 &  91 &  95 \\
LSTM $n$-gram, forward context encoding, straddling \verb+<S>+ & 13 &  87 &  91 \\\hline
\multicolumn{4}{|l|}{\bf{LSTM RNN, baseline}}\\\cline{1-1}
LSTM (medium setting) & $\infty$  & --- & 84 \\\hline
\end{tabular}
\caption{UPenn Treebank: perplexity values for neural network smoothed $n$-gram LM when straddling the sentence beginning boundary.}
\label{table:straddling-nn-ngram}
\end{table}

\subsection{One Billion Words Benchmark}
\label{one_bwds_exps}
In a second set of experiments we used the corpus in~\cite{Chelba:2014}, the same as in~\cite{DBLP:journals/corr/JozefowiczVSSW16}. For the baseline LSTM model we used the single machine implementation provided by~\cite{rafal:one-billion-wds}; the LSTM $n$-gram variant was implemented as a minor tweak on this codebase and is thus different from the one used in the UPenn Treebank experiments in Section~\ref{upenn_exps}.

We experimented with the LSTM configuration in Table~3 of~\cite{DBLP:journals/corr/JozefowiczVSSW16} for both baseline LSTM and $n$-gram variant, which are also the default settings in~\cite{rafal:one-billion-wds}: embedding and projection layer dimensionality was 128, one layer with state dimensionality of 2048. Training used Adagrad with gradient clipping by global norm (10.0) and droput (probability 0.1); back-propagation at the output soft-max layer is done using importance sampling as described in~\cite{DBLP:journals/corr/JozefowiczVSSW16} with a set 8192 ``negative'' samples. An additional set of experiments investigated the benefits of adding one more layer to both baseline and $n$-gram LSTM.

The results are presented in Tables~\ref{table:one-billion-words-results-not-straddling}-\ref{table:one-billion-words-results-straddling}; unlike the UPenn Treebank experiments, we did not tune the hyper-parameters for the $n$-gram LSTM and instead just used the same ones as for the LSTM baseline; as a result the perplexity values for the $n$-gram LSTM may be slightly suboptimal.

Similar to the UPenn Treebank experiments, we examined the effect of resetting state at sentence boundaries. As expected PPL dit not change significantly because the sentences in the training and test data were randomized, see~\cite{Chelba:2014}; in fact modeling the sentence independence explicitly is slightly beneficial.

We observe that on large amounts of data LSTM smoothing for short $n$-gram contexts does not provide significant advantages over classic back-off $n$-gram models. This may have implications for short-format text, e.g. voice search/query LMs. On the other hand, LSTM smoothing becomes very effective with long contexts ($n > 5$) approaching the fully recurrent LSTM model perplexity at about $n=13$.

Training times are significantly different between the LSTM baseline and the $n$-gram variant, with the latter being about an order of magnitude slower due to the fact that the LSTM state is recomputed and discarded for every new training sample.

\begin{table}[h!]
\centering
\begin{tabular}{|l|r|r|}
\hline
Model & Order & Test PPL\\\hline
\multicolumn{2}{|l|}{Training Target} & one-hot \\\hline\hline
\multicolumn{3}{|l|}{\bf{$n$-gram, baseline}}\\\cline{1-1}
Interpolated Kneser-Ney, back-off & 5 & 68 \\\hline
\multicolumn{3}{|l|}{\bf{LSTM RNN $n$-gram}}\\\cline{1-1}
LSTM $n$-gram, forward context encoding &  5 & 70 \\
LSTM $n$-gram, forward context encoding &  9 & 54 \\
LSTM $n$-gram, forward context encoding & 13 & 49 \\\hline
\multicolumn{3}{|l|}{\bf{LSTM RNN, baseline}}\\\cline{1-1}
LSTM & reset at \verb+<S>+  & 48 \\\hline
\multicolumn{3}{|l|}{\bf{2-layer LSTM RNN $n$-gram}}\\\cline{1-1}
LSTM $n$-gram, forward context encoding &  5 & 68 \\
LSTM $n$-gram, forward context encoding &  9 & 51 \\
LSTM $n$-gram, forward context encoding & 13 & 46 \\\hline
\multicolumn{3}{|l|}{\bf{2-layer LSTM RNN, baseline}}\\\cline{1-1}
LSTM & reset at \verb+<S>+  & 43 \\\hline
\end{tabular}
\caption{One Billion Words Benchmark: perplexity values for neural network smoothed $n$-gram LM when enforcing the sentence independence.}
\label{table:one-billion-words-results-not-straddling}
\end{table}
%

\begin{table}[h!]
\centering
\begin{tabular}{|l|r|r|}
\hline
Model & Order & Test PPL\\\hline
\multicolumn{2}{|l|}{Training Target} & one-hot \\\hline\hline
\multicolumn{3}{|l|}{\bf{LSTM RNN $n$-gram}}\\\cline{1-1}
LSTM $n$-gram, forward context encoding, straddling \verb+<S>+ &  5 & 70 \\
LSTM $n$-gram, forward context encoding, straddling \verb+<S>+ &  9 & 54 \\
LSTM $n$-gram, forward context encoding, straddling \verb+<S>+ & 13 & 49 \\\hline
\multicolumn{3}{|l|}{\bf{LSTM RNN, baseline}}\\\cline{1-1}
LSTM & $\infty$  & 49 \\\hline  
\multicolumn{3}{|l|}{\bf{2-layer LSTM RNN $n$-gram}}\\\cline{1-1}
LSTM $n$-gram, forward context encoding, straddling \verb+<S>+ &  5 & 68 \\
LSTM $n$-gram, forward context encoding, straddling \verb+<S>+ &  9 & 51 \\
LSTM $n$-gram, forward context encoding, straddling \verb+<S>+ & 13 & 46 \\\hline
\multicolumn{3}{|l|}{\bf{2-layer LSTM RNN, baseline}}\\\cline{1-1}
LSTM & $\infty$  & 43 \\\hline  
\end{tabular}
\caption{One Billion Words Benchmark: perplexity values for neural network smoothed $n$-gram LM when straddling the sentence beginning boundary.}
\label{table:one-billion-words-results-straddling}
\end{table}

%% file: conc.tex
\section{Conclusions and Future Work}
\label{conclusion}

We investigated the effective memory depth of (R)NN models by using them for word-level $n$-gram LM smoothing. The LSTM cell with dropout was by far the best (R)NN model for encoding the $n$-gram state.

When preserving the sentence independence assumption the LSTM $n$-gram matches the LSTM LM performance for $n=9$ and slightly outperforms it for $n=13$. When allowing dependencies across sentence boundaries, the LSTM $13$-gram almost matches the perplexity of the unlimited history LSTM LM.

We can thus conclude that the memory of LSTM LMs seems to be about 9-13 previous words which is not a trivial depth but not that large either.

Compared to standard $n$-gram smoothing methods LSTMs have excellent statistical properties: they improve with the $n$-gram order well beyond the point where Katz or Kneser-Ney back-off smoothing methods saturate, proving that they are able to extract richer features from the same context. Using multinomial targets in training is only slightly beneficial in this setting, although the advantage over one-hot diminishes with increasing $n$-gram order.

Experiments on the One Billion Words benchmark confirm that $n$-gram LSTMs can match the performance of fully recurrent LSTMs at larger amounts of data.

Building LSTM $n$-gram LMs is attractive due to the fact that the state in a $n$-gram LM can be succinctly represented on $4 \cdot (n-1)$ bytes storing the identity of the context words. This is in stark contrast with the state $H \in \mathbb{R}^h$ for an RNN LM, where $h=1024$ or higher, making the $n$-gram LM easier to use in decoders such as for ASR/SMT. The LM requests in the decoder can be batched, making the RNN LM operation more efficient on GPUs.

On the downside, the LSTM encoding for the $n$-gram context is discarded and cannot be re-used; caching it for frequent LM states is possible.